\documentclass{article}
\usepackage{spconf,amsmath,graphicx,multirow,multicol,amssymb,bbding,setspace}

\title{AN APPROACH TO MISPRONUNCIATION DETECTION AND DIAGNOSIS WITH ACOUSTIC, PHONETIC AND LINGUISTIC (APL) EMBEDDINGS
}
\name{Wenxuan Ye$^{1, *}$\thanks{*Work performed as intern in Microsoft}, Shaoguang Mao$^2$, Frank Soong$^2$, Wenshan Wu$^2$, Yan Xia$^2$, Jonathan Tien$^2$, Zhiyong Wu$^{1}$}

\address{
$^1$Shenzhen International Graduate School, Tsinghua University, Shenzhen, China\\
$^2$Microsoft Research Asia, Beijing, China\\
\texttt{ywx20@mails.tsinghua.edu.cn} \\ \texttt{\{shamao, frankkps, wenswu, yanxia, jtien\}@microsoft.com}, \\\texttt{zywu@se.cuhk.edu.hk}}
\begin{document}
\ninept
\maketitle
\begin{abstract}
Many mispronunciation detection and diagnosis (MD\&D) research approaches try to exploit both the acoustic and linguistic features as input. Yet the improvement of the performance is limited, partially due to the shortage of large amount annotated training data at the phoneme level. Phonetic embeddings, extracted from ASR models trained with huge amount of word level annotations, can serve as a good representation of the content of input speech, in a noise-robust and speaker-independent manner. These embeddings, when used as implicit phonetic supplementary information, can alleviate the data shortage of explicit phoneme annotations. We propose to utilize Acoustic, Phonetic and Linguistic (APL) embedding features jointly for building a more powerful MD\&D system. Experimental results obtained on the L2-ARCTIC database show the proposed approach outperforms the baseline by 9.93\%, 10.13\% and 6.17\% on the detection accuracy, diagnosis error rate and the F-measure, respectively.
\end{abstract}
\begin{figure*}[!t]
\centering
    {\includegraphics[width=1.\linewidth]{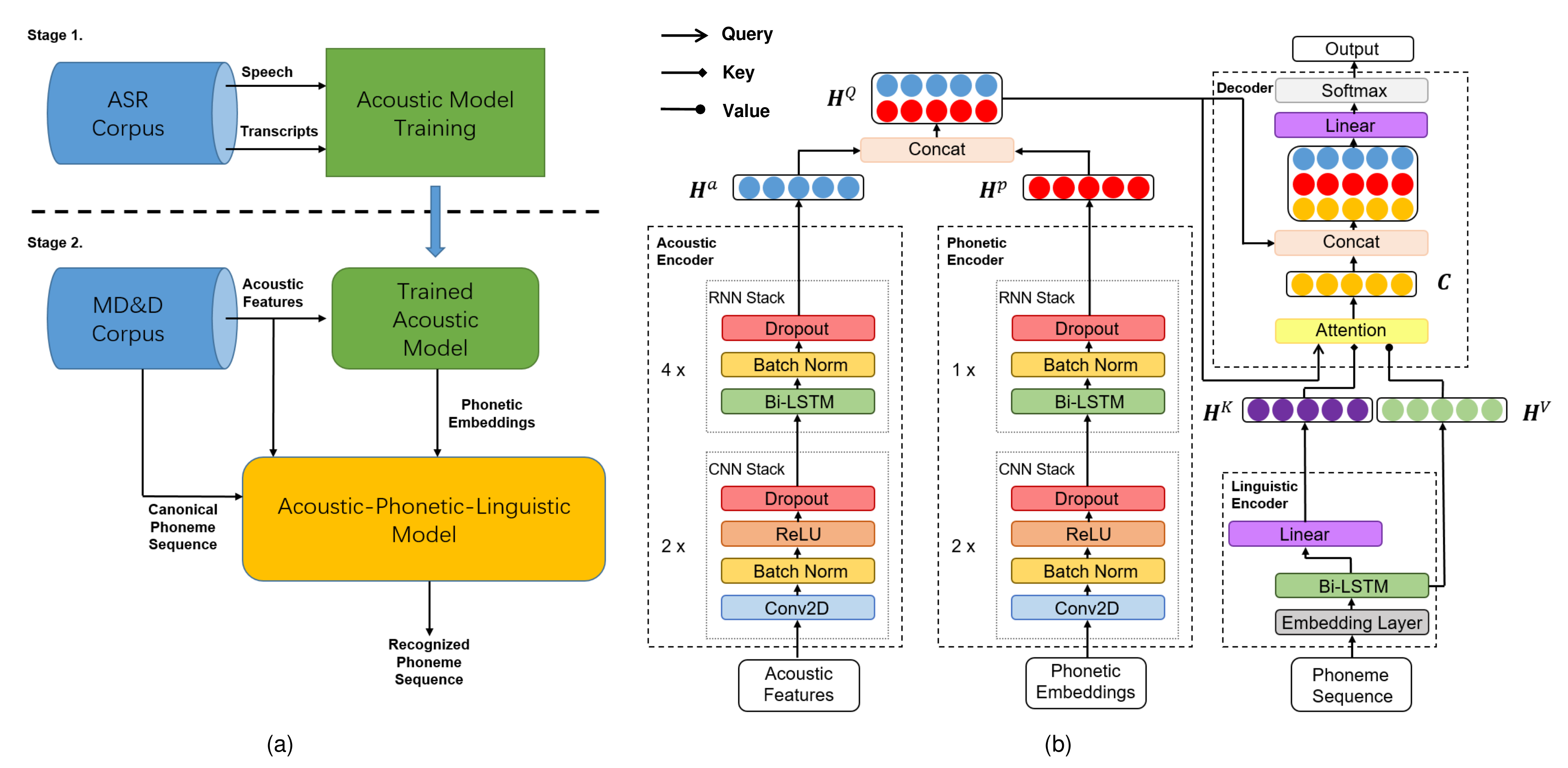}}
    \caption{Illustration of the proposed approach to Mispronunciation Detection and Diagnosis with acoustic features, phonetic embeddings and linguistic embeddings (APL) (a) Training flow; (b) APL model architecture.}
    \label{model_structure}
\end{figure*}

\section{Introduction}
\label{sec:intro}

The development of Computer-aided Pronunciation Training(CAPT) system empowers language learners a convenient way to practice their pronunciations\cite{importance_1,importance_2,importance_3}, especially for those who have little access to professional teachers. 

Mispronunciation Detection and Diagnosis (MD\&D) is a key part of CAPT and 
several methods have been proposed to tackle it. 
Goodness of Pronunciation (GOP)\cite{gop}, developed by Witt and Young, computes scores based on log-posterior probability from acoustic models and then detects mispronunciation with phone-dependent thresholds. Even though these kinds of approaches provide scores for mispronunciation detection\cite{extended_gop, transfer_logistic, context_aware}, they cannot provide sufficient diagnosis information for pronunciation correction.
To better obtain diagnosis information, Extended Recognition Network (ERN)\cite{ern1, ern2, ern3} extends the decoding stage of Automatic Speech Recognition (ASR) by modeling pre-defined context-dependent phonological rules. 
However, ERN fails to deal with the mispronunciation patterns which are absent in training data or manual rules. 
Additionally, when too many phonological rules are included in ERN, recognition accuracy may be affected, thus leading to unreliable MD\&D feedbacks.

Moreover, since the above-mentioned approaches inevitably involve multiple training stage, complicated manual designs for each stage are required and the performances are sensitive to the precision of each stage. Recently, a number of researches have proposed end-to-end models for phone-level MD\&D. CNN-RNN-CTC\cite{cnn_rnn_ctc}, a phoneme recognizer directly mapping the raw acoustic features to corresponding pronounced phone sequences with the help of connectionist temporal classification (CTC) loss function\cite{ctc}, shows potentials of end-to-end approaches. Whereas,  end-to-end training depends on adequate well-labeled data.  Imagining the challenges of phoneme labeling, labeling a unit in 40-60ms, and the non-standard pronunciations from second language (L2) speakers, a large-scale and labeled second language learner's speech are hard to collect.

In MD\&D scenarios, the canonical phoneme sequences are available and L2 learners' pronunciation will be checked by contrast with the canonical phoneme sequences. Therefore, linguistic information from canonical text can be integrated into models to promote MD\&D performance. SED-MDD\cite{sedmdd} leverages attention mechanism to incorporate acoustic features and corresponding canonical character sequence. And \cite{fulltext} aligns acoustic features with canonical phoneme sequence from a sentence encoder to decode the pronounced phoneme sequence. Even if linguistic embedding brings extra information for MD\&D, the scarcity of data to train a capable acoustic encoder is still a challenge to overcome.

Although phoneme-level annotations are hard to collect, word-level transcripts for ASR training are relatively adequate and easy to access. Compared with acoustic features, which may be easily influenced by noises or the speakers variances, the phonetic embedding, extracted from a well-trained ASR model, may represent phonetic information in a noise-robust and speaker-independent manner. Specifically, the phonetic embedding could be the output from an ASR model like phonetic posteriorgrams (PPGs) or bottleneck features (BNFs) from hidden layers. Considering that ASR models are tailored for word recognition, which is subtly different from phoneme recognition in MD\&D, phonetic embedding should be a supplement, not a substitute for acoustic features. Thus, both acoustic features and phonetic embedding should be taken for MD\&D phoneme recognizer training.

Riding on these ideas, we propose a phoneme-level MD\&D framework which employs acoustic embedding (acoustic features), phonetic embedding and linguistic embedding (canonical phoneme sequence) as inputs to predict the pronounced phoneme sequence. Compared with the previous works, the proposed method innovatively adopts the information distilled from well-trained L1 ASR models to resolve MD\&D tasks.  This information contains relatively robust phonetic distribution information and is a supplement for raw acoustic and linguistic features.  The proposal utilizes enormous L1 ASR datasets to relieve the data scarcity of MD\&D. Meanwhile, experiments conducted on the latest version of L2-ARCTIC corpus\cite{l2_arctic} verify the proposal's efficiency.

\section{PROPOSED METHOD}
\label{sec:method}

The proposed model is shown in Fig.1 (b) \cite{fulltext}. It consists of an \textbf{a}coustic encoder, a \textbf{p}honetic embedding encoder, a \textbf{l}inguistic encoder, and a decoder with attention, so it is correspondingly called APL. As the illustration shows, the model takes  Mel-filter banks (fbanks), phonetic embeddings extracted from pre-trained acoustic model and canonical phoneme sequence as input respectively and outputs recognized phoneme sequence. The model is jointly trained with CTC loss\cite{ctc}.

\subsection{Acoustic Encoder}
\label{ssec: audio_encoder}
The input of acoustic encoder $E_a$ is an 81-dim acoustic feature vector $\boldsymbol{X}=[\boldsymbol{x}_1, ..., \boldsymbol{x}_t, ..., \boldsymbol{x}_T]$ (80-dim fbanks and 1-dim energy), where $T$ stands for the number of frames of input speech. $E_a$ consists of two convolution neural network (CNN) stacks and four recurrent neural network (RNN) stacks in order. In details, the CNN stack starts with a 2D convolution layer, followed by a batch normalization layer, one ReLU activation function and a dropout layer. The RNN stack includes a bi-directional LSTM layer, a batch normalization layer and a dropout layer.

High-level acoustic representations are obtained by $E_a$ from the input $\boldsymbol{X}$:
\begin{equation}
    \boldsymbol{H}^{a}=E_a(\boldsymbol{X}) \label{eq_audio_hidden}
\end{equation}
where $\boldsymbol{H}^{a}=[\boldsymbol{h}^{a}_1, ..., \boldsymbol{h}^{a}_{t^\prime}, ..., \boldsymbol{h}^{a}_{T^\prime}]$ is the encoded acoustic features with $T^\prime$ frames.

\subsection{Phonetic Encoder}
\label{ssec: ppg_encoder}
The model takes phonetic embeddings $\boldsymbol{P}=[\boldsymbol{p}_1, ..., \boldsymbol{p}_t, ..., \boldsymbol{p}_T]$ as input, which are extracted by pre-trained ASR models and have an identical number of frames with acoustic features $\boldsymbol{X}$. Before integrated with other inputs, the phonetic embeddings are fed into an encoder $E_{p}$ to derive its representations $\boldsymbol{H}^{p}=[\boldsymbol{h}^{p}_1, ..., \boldsymbol{h}^{p}_{t^\prime}, \boldsymbol{h}^{p}_{T^\prime}]$:
\begin{equation}
    \boldsymbol{H}^{p}=E_{p}(\boldsymbol{P}) \label{eq_ppg_hidden}
\end{equation}
Similar to the acoustic encoder, the phonetic embedding encoder is also composed of CNN stacks and RNN stacks. 

To ensure $\boldsymbol{H}^{p}$ and $\boldsymbol{H}^{a}$ have the same time resolution, the CNN stacks in phonetic encoder are exactly the same as audio encoder. Since these embeddings are relatively high-level representations compared with raw acoustic features, only one RNN stack is in the encoder $E_{p}$.

\subsection{Linguistic Encoder}
\label{ssec:linguistic_encoder}
Considering the characteristic of MD\&D task, where the canonical phoneme sequences are available for mispronunciation check\cite{multi_dist}, a linguistic encoder $E_l$ serves the purpose of extracting linguistic representations of a given utterance from its canonical phoneme sequence $\boldsymbol{s}=[s_1, ..., s_n, ..., s_N]$ with $N$ phonemes:
\begin{equation}
    \boldsymbol{H}^K,\boldsymbol{H}^V=E_l(\boldsymbol{s})
\end{equation}
$\boldsymbol{H}^K=[\boldsymbol{h}^K_1, ..., \boldsymbol{h}^K_n, ..., \boldsymbol{h}^K_N]$ and $\boldsymbol{H}^V=[\boldsymbol{h}^V_1, ..., \boldsymbol{h}^V_n, ..., \boldsymbol{h}^V_N]$ are sequential embeddings to be used as keys and values in the decoder. 

\subsection{Decoder}
A decoder with attention mechanism is utilized to integrate information from acoustic, phonetic and linguistic (APL) encoders. The $\boldsymbol{H}^{a}$ and $\boldsymbol{H}^{p}$ are concatenated together to compose the query $\boldsymbol{H}^Q$ in attention, representing the extracted acoustic features. For a given frame $t^{\prime}$ we have
\begin{equation}
    \boldsymbol{h}_{t^\prime}^Q=[\boldsymbol{h}_{t^\prime}^{a};\boldsymbol{h}_{t^\prime}^{p}] \label{eq_query_hidden}
\end{equation}
where
$[.;.]$ denotes the concatenation of two vectors. Then the normalized attention weight between frame $\boldsymbol{h}_{t^\prime}^Q$ in $\boldsymbol{H}^Q$ and $\boldsymbol{h}_n^K$ in $\boldsymbol{H}^K$ can be computed by
\begin{equation}
        \alpha_{t^\prime,n}=\frac{\exp{(\boldsymbol{h}_{t^\prime}^Q {\boldsymbol{h}_n^K}^T})}{\sum_{n=1}^N \exp{(\boldsymbol{h}_{t^\prime}^Q {\boldsymbol{h}_n^K}^T})}
\end{equation}

Further, the context vector $\boldsymbol{c}_{t^\prime}$ at frame $t^\prime$ obtained by aligning the acoustic features with linguistic features is given by
\begin{equation}
    \boldsymbol{c}_{t^\prime}=\sum_{n}^N \alpha_{t^\prime,n} \boldsymbol{h}_n^V
\end{equation}

Note that the context vector $\boldsymbol{c}_{t^\prime}$ is the weighted average of $\boldsymbol{h}_n^V$, which comes from linguistic representations of a given sentence. The information may be inadequate to represent those mispronounced phonemes that are absent from the canonical phoneme sequence $\boldsymbol{s}$. Hence, in the output layer, the frame-wise probability $\boldsymbol{y}_{t^\prime}$ is computed from both $\boldsymbol{c}_{t^\prime}$ and $\boldsymbol{h}_{t^\prime}^Q$:
\begin{equation}
    \boldsymbol{y}_{t^\prime}=softmax(\boldsymbol{W}[\boldsymbol{c}_{t^\prime};\boldsymbol{h}_{t^\prime}^Q]+b)
\end{equation}
where $\boldsymbol{W}$ and $b$ are weight matrix and bias of output layer. Finally, the recognized phoneme sequence is obtained by beam-search on $\boldsymbol{y}_{t^\prime}$.

\section{EXPERIMENTS}
\label{sec:experiments}

\subsection{Datasets}
\label{ssec:datasets}
Our experiments are conducted on TIMIT\cite{timit} and L2-ARCTIC (V5.0)\cite{l2_arctic} corpus. TIMIT contains recordings of 630 US native speakers and L2-ARCTIC includes recordings of 24 non-native speakers whose mother tongues are Hindi, Korean, Mandarin, Spanish, Arabic and Vietnamese. 
The speakers from each language contain recordings of two males and two females. 
Both of the two corpora are publicly available and include phoneme-level annotations.

We follow the setting in \cite{phone_map} to map the 61-phone set in TIMIT and 48-phone set in L2-ARCTIC to the 39-phone set. Moreover, the L2-ARCTIC corpus contains 28 additional phonemes with foreign accents, marked with a “deviation” symbol “*”. And if a perceived phoneme was hard to judge, it would be annotated as “err” in L2-ARCTIC. These 29 special annotations are treated as independent classes along with 39 standard phones in our experiments. More details are discussed on the websites \footnote{https://psi.engr.tamu.edu/l2-arctic-corpus-docs/}.

The data split is shown in Table.\ref{table_split}. To ensure all classes in dev/test set to be included in the training set, the speaker splits of L2-ARCTIC are as: dev set (EBVS, THV, TNI, BWC, YDCK, YBAA), test set (NJS, HQTV, SVBI, NCC, YKWK, ZHAA), training set (all other speakers).

\begin{center}
\begin{table}[ht]
\centering
\caption{Details of dataset used in the experiments}
\label{table_split}
\begin{tabular}{|l|c|c|c|c|}
\hline
\multirow{2}{*}{} & TIMIT      & \multicolumn{3}{c|}{L2-ARCTIC}     \\ \cline{2-5} 
                  & \multicolumn{2}{c|}{Training} & dev     & test     \\ \hline
Speakers          & 630        & 12            & 6       & 6        \\ \hline
Utterances        & 6300       & 1800          & 897     & 900      \\ \hline
\end{tabular}

\end{table}
\end{center}
 
\begin{table*}[]
\centering
\small
\caption{Results of phoneme recognition and MD\&D}
\label{table_pr}
\begin{tabular}{|l|c|c|c|c|c|c|c|c|c|}
\hline
\multicolumn{1}{|c|}{\multirow{2}{*}{Models}} & \multicolumn{2}{c|}{Phoneme Recognition} & \multicolumn{7}{c|}{Mispronunciation Detection and Diagnosis}                                                                                                                                                           \\ \cline{2-10} 
\multicolumn{1}{|c|}{}                        & Correctness         & Accuracy           & FRR             & FAR              & \begin{tabular}[c]{@{}c@{}}Detection\\ Rate\end{tabular} & \begin{tabular}[c]{@{}c@{}}Diagnosis\\ Error Rate\end{tabular} & Recall           & Precision        & F-measure        \\ \hline
Baseline-1                                    & 71.95\%             & 70.25\%            & 24.45\%         & \textbf{24.81\%} & 74.81\%                                                  & 44.14\%                                                        & \textbf{74.19\%} & 34.70\%          & 47.28\%          \\ \hline
Baseline-2 (AL)                                    & 73.21\%             & 71.22\%            & 24.19\%         & 27.60\%          & 75.66\%                                                  & 44.07\%                                                        & 72.40\%          & 35.29\%          & 47.45\%          \\ \hline
PL-1                                          & 82.18\%             & 80.47\%            & 12.12\%         & 41.83\%          & 83.31\%                                                  & 37.67\%                                                        & 58.17\%          & 46.66\%          & 51.78\%          \\ \hline
PL-2                                          & 81.11\%             & 79.56\%            & 13.55\%         & 39.43\%          & 82.63\%                                                  & 37.64\%                                                        & 60.57\%          & 44.89\%          & 51.57\%          \\ \hline
APL-1                                         & 81.49\%             & 79.65\%            & 13.57\%         & 38.01\%          & 82.84\%                                                  & 37.89\%                                                        & 61.99\%          & 45.43\%          & 52.44\%          \\ \hline
APL-2                                         & \textbf{84.58\%}    & \textbf{83.04\%}   & \textbf{8.88\%} & 45.51\%          & \textbf{85.59\%}                                         & \textbf{33.94\%}                                               & 54.49\%          & \textbf{52.79\%} & \textbf{53.62\%} \\ \hline
\end{tabular}
\end{table*}

\subsection{Acoustic Models}
\label{ssec:am}
Due to the quality of the phonetic embeddings mentioned in 2.3 depends on data quality and quantity for ASR training, two acoustic models are involved in the experiments to verify the proposed method's robustness. AM1 is an acoustic model proposed by \cite{any_to_many_vc} trained on LibriSpeech\cite{librispeech} corpus, which produces a 144-dim frame-wise bottleneck features from raw acoustic input. The AM2 is trained on Microsoft EN* dataset, which contains speech of more than 100k hours from English speakers across the world, and derives a 41-dim frame-level PPGs.

\subsection{Experimental Setups}
\label{ssec:exp_setup}
Six models are implemented for comparisons:
\begin{itemize}
\item
\textbf{Baseline-1}: CNN-RNN-CTC \cite{cnn_rnn_ctc}; 
\item
\textbf{Baseline-2 (AL)} \cite{fulltext}: taking acoustic features (fbanks) and linguistic embeddings (canonical phoneme sequences) as input, CTC as loss;  
\item
\textbf{PL-1}: taking \textbf{p}honetic embedding from AM1 and \textbf{l}inguistic embedding as input, CTC as loss; 
\item
\textbf{PL-2}: taking \textbf{p}honetic embedding from AM2 and \textbf{l}inguistic embedding as input, CTC as loss; 
\item
\textbf{APL-1}: taking \textbf{a}coustic features (fbanks), \textbf{p}honetic embedding from AM1 and \textbf{l}inguistic embedding as input, CTC as loss; 
\item
\textbf{APL-2}: taking \textbf{a}coustic features (fbanks), \textbf{p}honetic embedding from AM2 and \textbf{l}inguistic embedding as input, CTC as loss; 
\end{itemize}

PL-1 and PL-2 are conducted to investigate the differences between taking phonetic embeddings extracted from well-trained ASR model or raw acoustic features as input. APL-1 and APL-2 are implemented to verify the efficiency of taking acoustic, phonetic and linguistic embeddings together as input for MD\&D. Besides, when comparing PL-1 and APL-1 with Model PL-2 and APL-2 respectively, the influence of L1 ASR acoustic model training can be observed.

The parameters of acoustic encoder, phonetic encoder, linguistic encoder and decoder are the same as baselines. All audios are in a 16k sampling rate. Fbanks and phonetic embedding are computed with 10ms shift. All models are trained with a batch size of 64 utterances and 200 maximum epochs.

\begin{figure}[!t]
\centering
    {\includegraphics[width=1.\linewidth]{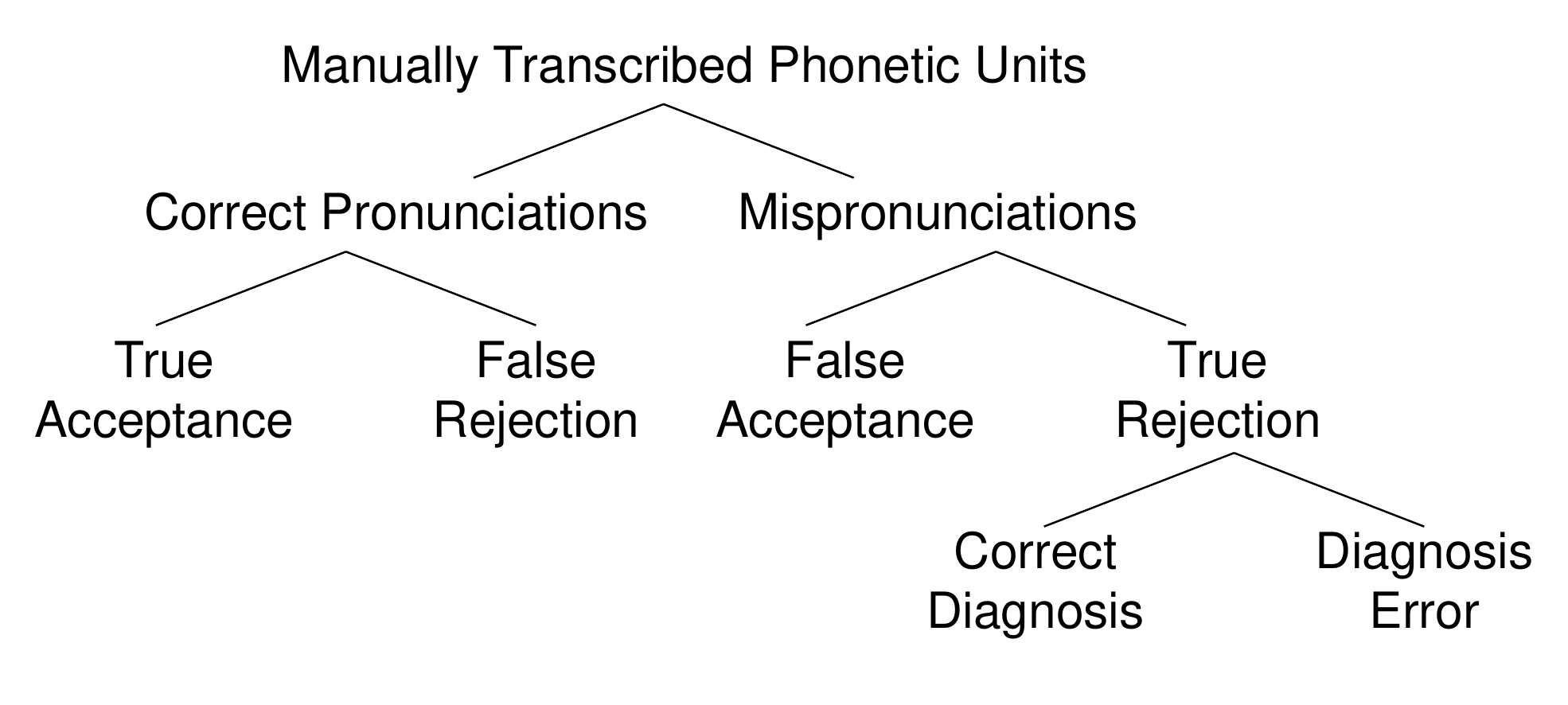}}
    \caption{Hierarchical evaluation structure for mispronunciation detection and diagnosis \cite{multi_dist}.}
    \label{hierarchical}
\end{figure}

\subsection{Phoneme Recognition}
\label{ssec:pr}
All models in 3.3 output recognized phoneme sequences. The recognized results are aligned with human annotations based on editing distance. The metrics are computed as (\ref{correctness}) and (\ref{accuracy}), where \textit{I} indicates insertions, \textit{D} indicates deletions, \textit{S} indicates substitutions, and \textit{N} indicates number of all phonetic units.
\begin{equation}
    Correctness=\frac{N-S-D}{N} \label{correctness}
\end{equation}

\begin{equation}
    Accuracy=\frac{N-S-D-I}{N} \label{accuracy}
\end{equation}
As shown in Table \ref{table_pr}, with the additional linguistic information, Baseline-2 slightly outperforms Baseline-1.

Comparing Baseline-2 with PL-1 and PL-2, we can find that taking phonetic embedding extracted from an ASR model as input significantly performs better than taking acoustic features. Because the phonetic embeddings extracted from the ASR model contain phonetic distribution learned from rich resources. Under an inadequate data training scenario, it is efficient to leverage models in rich resource cases to represent the articulated content of input speech. 

When the acoustic feature is appended, APL-1 and APL-2 further surpass the PL-1, PL-2 and Baseline 2. Especially the APL-2 achieves state-of-the-art. The results verify our assumption that acoustic features, phonetic embeddings and linguistic embeddings are all necessary for L2 phoneme recognition. When they are combined together, the joint embedding learning shows its potential. 

\subsection{Mispronunciation Detection and Diagnosis}
\label{ssec:mdd}
The hierarchical evaluation structure in \cite{multi_dist} is adopted to measure the MD\&D performance, as shown in Fig.\ref{hierarchical}. The correct detections include true acceptance (TA) and true rejection (TR), while the incorrect detections are false rejection (FR) and false acceptance (FA). And those cases in TR are further split into correct diagnosis (CD) and diagnosis error (DE). Then the metrics for mispronunciation detection are calculated follow (\ref{FRR}) - (\ref{F-measure}).
\begin{equation}
    FRR=\frac{FR}{TA+FR} \label{FRR}
\end{equation}
\begin{equation}
    FAR=\frac{FA}{FA+TR} \label{FAR}
\end{equation}
\begin{equation}
    Detection\:Accuracy=\frac{TR+TA}{TR+TA+FR+FA} \label{Detection_Rate}
\end{equation}
\begin{equation}
    Precision=\frac{TR}{TR+FR} \label{Precision}
\end{equation}
\begin{equation}
    Recall=\frac{TR}{TR+FA}=1-FAR \label{Recall}
\end{equation}
\begin{equation}
    F-measure=2\frac{Precision*Recall}{Precision+Recall} \label{F-measure}
\end{equation}

The Diagnosis Error Rate (DER) for mispronunciation diagnosis is calculated as (\ref{DER}): 
\begin{equation}
    Diagnosis\:Error\:Rate=\frac{DE}{DE+CD} \label{DER}
\end{equation}

As presented in Table \ref{table_pr}, the best detection accuracy (85.59\%), diagnosis error rate (33.94\%) and F-measure (53.62\%) occur when acoustic features, phonetic embeddings and linguistic information are given together as input. The APL-2 gains 9.93\%, 10.13\% and 6.17\% improvements on detection accuracy, diagnosis error rate and F-measure against Baseline-2 respectively. The contrast experimentally verifies the efficiency of added phonetic embeddings.

 It is worth mentioning that the performance of PL-1 is very close to PL-2, but when acoustic features are appended, APL-2 significantly outperforms APL-1. We assume that AM2 is a deeper model trained on a larger dataset (100k v.s. 960 hours), so the phonetic representations by AM2 are well noise-tolerant and speaker-normalized but may lose some useful information for MD\&D. When the acoustic features are provided, the encoders obtain better representations to align with the linguistic embedding, thus producing a more accurate output. More supplementary analysis can be found on the website\footnote{ https://thuhcsi.github.io/icassp2022-MDD-APL/}.

\section{CONCLUSION}
\label{sec:conclusion}
We propose a model which incorporates acoustic, phonetic and linguistic (APL) embedding features for improving MD\&D performance. The phonetic embeddings are extracted from a well-trained, speaker-independent, noise-robust ASR model without using phoneme-level annotated data. With the combination of acoustic features, phonetic embedding and linguistic embeddings derived from the canonical phoneme sequence, the APL approach can achieve significant improvements on phoneme recognition and MD\&D performance. Testing results on the L2-ARCTIC database show that the proposed approach is effective for improving the detection accuracy, diagnosis error rate and F-measure over the baseline system by 9.93\%, 10.13\% and 6.17\%, respectively.

\bibliographystyle{IEEEbib}
\bibliography{strings,refs}

\end{document}